\documentclass[12pt,a4paper]{article}

\usepackage{graphicx}
\usepackage{amsmath}
\usepackage{amssymb}
\usepackage[font=small]{caption}
\usepackage{fullpage}
\usepackage[numbers]{natbib}
\usepackage{color}

\usepackage{setspace}

\usepackage[T1]{fontenc}
\usepackage[utf8]{inputenc}
\usepackage{authblk}
\usepackage{url}

\usepackage{soul,color}
\usepackage{float}
\usepackage{multirow}

\title{Comparative study of LSA vs Word2vec embeddings in small corpora: a case study in dreams database}
\author[1]{Edgar Altszyler \thanks{Corresponding author: ealtszyler@dc.uba.ar}}
\author[2]{Mariano Sigman}
\author[3]{Sidarta Ribeiro}
\author[1]{Diego Fern\'andez Slezak}

\affil[1]{Laboratorio de Inteligencia Artificial Aplicada, Depto. de Computaci\'on, Universidad de Buenos Aires - CONICET.}
\affil[2]{Universidad Torcuato Di Tella - CONICET.}
\affil[3]{Instituto do C\'erebro, Universidade Federal do Rio Grande do Norte, Natal, Brazil.}

\date{\vspace{-5ex}}

\begin{document}
\maketitle

\begin{abstract}
Word embeddings have been extensively studied in large text datasets. However, only a few studies analyze semantic representations of small corpora, particularly relevant in single-person text production studies. 
In the present paper, we compare Skip-gram and LSA capabilities in this scenario, and we test both techniques to extract relevant semantic patterns in single-series dreams reports. 
LSA showed better performance than Skip-gram in small size training corpus in two semantic tests.
As a study case, we show that LSA can capture relevant words associations in dream reports series, even in cases of small number of dreams or low-frequency words. We propose that LSA can be used to explore words associations in dreams reports, which could bring new insight into this classic research area of psychology

\end{abstract}

\section{Introduction}
Corpus-based semantic representations (i.e. embeddings) exploits statistical properties of textual structure to embed words in a vectorial space. 
In this space, terms with similar meanings tend to be located close to each other. 
These methods rely in the idea that words with similar meanings tend to occur in similar contexts \cite{Harris1954}. 
This proposition is called \textit{distributional hypothesis} and provides a practical framework to understand and compute semantic relationship between words. 

Word embeddings has been used in a wide variety of applications such as sentiment analysis \cite{Socher2012}, psychiatry \cite{Bedi2015}, psychology \cite{Sagi2013,elias2009scale}, philology \cite{Diuk2012}, cognitive science \cite{landauer2007} and social science \cite{Carrillo2015,Kulkarni2015}.

Latent Semantic Analysis (LSA) \cite{deerwester1990indexing,Landauer1997,Hu2007}, is one of the most used methods for word meaning representation.
LSA takes as input a training corpus, i.e. a collection of documents. 
A word by document co-occurrence matrix is constructed. 
Typically, normalization is applied to reduce the weight of uninformative high-frequency words in the words-documents matrix  
\cite{Dumais1991}. 
Finally, a dimensionality reduction is implemented by a truncated \textit{Singular Value Decomposition}, SVD, which projects every word in a subspace of a predefined number of dimensions. Once the vectorial representation of words is obtained, the semantic similarity between two terms is typically computed by the cosine of the angle between them.  

More recently, neural-network language embeddings have received an increasing attention \cite{Collobert2008,Mikolov2013a}, leaving aside classical word representation methods such as LSA. In particular, Word2vec models \cite{Mikolov2013a,Mikolov2013b} have become especially popular in embeddings generation. 

Word2vec consists of two neural network language models, Continuous Bag of Words (CBOW) and Skip-gram. In both models, a window of predefined length is moved along the corpus, and in each step the network is trained with the words inside the window. 
Whereas the CBOW model is trained to predict the word in the center of the window based on the surrounding words, the Skip-gram model is trained to predict the contexts based on the central word. 
Once the neural network has been trained, the learned linear transformation in the hidden layer is taken as the word representation. In the present paper, we use Skip-gram model, which shows better performance in \cite{Mikolov2013b} semantic task.

An intrinsic difference between LSA and Word2vec is that while LSA is a counter-based model, Word2vec is a prediction-based model. Although prediction-based models have strongly increased in popularity, it is not clear whether they outperform classical counter-based models \cite{Baroni2014,Levy2014,Levy2015}. 

In particular, Word2vec methods have a distinct advantage in handling large datasets, since they do not consume as much memory as some classic methods like LSA and, as part of the Big Data revolution, Word2vec has been trained with large datasets of about billions of tokens. 
However, often in several problems of natural and social sciences one has to form semantic embeddings based on scarce data. 
For example, when analyzing the semantic map of a psychiatric patient or tracking the semantic network growth in children's writing. 
Moreover, this kind of approach is also relevant in sociological and linguistic research, in which linguistic patterns in word meaning networks are tracked along the time line, and small time chunks are needed \cite{Kulkarni2015,Sotiropoulos2015}.

Which is the best method when only small amounts of data are available?
In the present paper we investigate this, on the working hypothesis that Word2vec will produce very low quality embeddings when trained with small corpus, as it is a prediction-based model and it would need lot of training data in order to fit its high number or parameters. Here we examine this hypothesis, investigating the optimality of different methods to achieve reliable semantic mappings when only medium to small corpora  are available for training. In these conditions, we compare Word2vec performances with LSA in a semantic categorization test, in which the capabilities of the model to represent semantic categories (such as, drinks, countries, tools and clothes) is measured. 

Then we examine the performance of these models in a real-life and challenging problem based on relatively short texts: analyzing and disambiguating the content of dreams. Dream content show gender and cultural differences, consistency over time of the dreams content, and concordance of dreaming features (such as activity and emotions) with waking-life experiences \cite{bell2011personality,Domhoff2002,Domhoff2008}. Also, there is evidence of change in dreams contents after drug treatment \cite{kirschner1999} and shifts in content patterns in people with psychiatric disorders \cite{domhoff2000methods}. 

Most of the newest dreams content analysis methods are based on frequency word-counting of predefined categories in dreams reports \cite{Domhoff2008}. A well known limitation of this approach is the impossibility of identifying the meaning of the counted words, which is determined by the context in which they appears. For example, the occurrence of the word \textit{fall} in a dream report may be used in different contexts, such as, \textit{falling} from a cliff, teeth \textit{falling} out or \textit{falling} sick. In this context, we will test the capabilities of LSA and Word2vec on identify patterns in the usage of words among subjects. In particular, we set to analyze the semantic neighborhood of the word \textit{run} present in the dreams reports of the different subjects. We have chosen this word because its frequency in dreams and the variety of contexts where it can be used. For example, \emph{run} may be associated to sports activities or with chase/escape situations, which is reported to be one of the most typical dreams \cite{Nielsen2003,Griffith1958}.

Here we specifically analyze the capabilities of both models to identify word associations in dreams reports. 
We believe that word embeddings can bring new insights in dreams content analysis.  
On the other hand, we claim that LSA would be more appropriate in small-size corpus and should outperform Word2Vec performance in this context.

\section{Methods}
\subsection{Semantic representations}
Both, LSA and Word2vec semantic representations were generated with the Gensim Python library \cite{rehurek_lrec}.
In LSA implementation, a tf-idf transformation was applied before the truncated Singular Value Decomposition. LSA's representation dimensionality were tuned in order to maximize its performance in each case. In Word2vec (Skip-gram) implementations no minimum frequency threshold were used, and the window size, the number of negative samples and the representation dimensionality were tuned to maximize the performance. All other Skip-gram parameters were set to default Gensim values.

Given a vectorial representation, the semantic similarity ($S$) of two words was calculated using the cosine similarity measure between their respective vectorial representation ($\mathbf{v_1}$,$\mathbf{v_2}$),
\begin{equation}
S(\mathbf{v_1},\mathbf{v_2})=cos(\mathbf{v_1},\mathbf{v_2})=\frac{\mathbf{v_1}.\mathbf{v_2}}{\|\mathbf{v_1}\|.\|\mathbf{v_2}\|}
\end{equation}

The semantic distances between two words $d(\mathbf{v_1},\mathbf{v_2})$ was calculated as 1 minus the semantic similarity ( $d(\mathbf{v_1},\mathbf{v_2}) = 1- S(\mathbf{v_1},\mathbf{v_2}) $).

\subsection{Semantic tests}
To compare LSA and Skip-gram semantic representation quality, we perform two tests in two different corpora (TASA and UkWaC): (1) a semantic categorization test and (2) a word-pairs similarity test. For each test, we studied how the performance of LSA and Skip-gram embeddings depend on the corpus size.  To do this, we take 6 nested sub-samples of the training corpora, in which documents where progressively eliminated, following \cite{Bullinaria2007,bullinaria2012extracting}. In both cases, the minimum
sub-corpus size contains only 600 documents. When any of the test words did not appear at least once in a sub-corpus, a random document was replaced with one of the discarded ones.

\subsubsection{Semantic categorization test}\label{method_categs}
In this test we measured the capabilities of the model to represent the semantic categories \cite{Patel1997,Bullinaria2007} (such as, drinks, countries, tools and clothes). 
The test is composed by 53 categories with 10 words each. 
In order to measure how well the word $i$ is grouped vis-\`{a}-vis the other words in its semantic category we used the Silhouette Coefficients, $s(i)$ \cite{Rousseeuw1987}, 
\begin{equation}
s(i) = \frac{b(i) - a(i)}{\max\{a(i),b(i)\}},
\end{equation}
where $a(i)$ is the mean distance of word $i$ with all other words within the same category, and $b(i)$ is the minimum mean distance of word $i$ to any words within another category (i.e. the mean distance to the neighboring category). 
In other words, Silhouette Coefficients measure how close a word is to other words within the same category compared to words of the closest category.
The Silhouette Score is computed as the mean value of all Silhouette Coefficients. 
The score takes values between -1 and 1, higher values reporting localized categories with larger distances between categories, representing better clustering. 

\subsubsection{Word-pairs similarity test}\label{method_wordsim}
This test measures the capabilities of the model to capture semantic similarity between concepts. We
used the well established WordSim353 test collection \cite{finkelstein2001placing}, which consist of 353 word-pairs (such as Maradona-football or physics-chemistry) associated with a mean human-assigned similarity score. Each word-pair is rated on a scale ranging from 0 (highly dissimilar words) to 10 (highly similar words).
The evaluation score is computed as the Spearman correlation between the human scores and the model semantic similarities.

\subsection{Case study: Semantic association in dreams reports}\label{method_dreams}
In this case study, we analyze the capabilities of the models to capture semantic word associations, testing whether the models embeddings can capture the semantic neighborhood of a target word in single subject's dream series ( a collection of dream reports written by the same person). In particular, we selected the word \textit{run} as the target word, and we focused on the detection of its distance to escape/chase contexts. The rank distance of a given word ``w'' with respect to \textit{run} was measured as the rank of ``w'' among the cosine similarity between \textit{run} and all other words in the vocabulary. For example, if a word has a rank of \textit{20}, it means that, among all words in the vocabulary, it is the 20th closest word using a cosine similarity metric. Finally, we will define the rank distance of escape/chase concepts as the minimum value within the ranks of the words \textit{escape*} and \textit{chase*} \footnote{The asterisk (*) refers to a word in all its forms, i.e. escape* stands for escape, escapes, escaping and escaped.}.

For each dream series, two independent annotators read all the dreams in which the word \textit{run} appears, and labeled whether they refer to an escape/chase situation or not. Escape/chase situations were defined as those in which (1) someone is being chased or is under the impression of being chased or (2) someone is escaping from a real or imaginary threat. Also, for a \textit{run} frame to be counted as an escape/chase context, it must be associated to a negative emotional valence, thus discarding, for instance, escape/chase situations related to games or sports.  
With these criteria, for each dream series the annotators calculate the fraction of times the word \textit{run} appears in an escape/chase context, obtaining a Pearson correlation coefficient of 0.98. We use the average of the annotators measurement as the ground truth, and we will refer to this values as the \textit{escape/chase fraction}. 

We used the \textit{escape/chase fraction} as a ground truth to test the embeddings quality. 
Good representations should produce low rank distance in series with high \emph{escape/chase fraction} and high rank distance in series with low \emph{escape/chase fraction}. 
Thus, not only do we expect negative correlations between the escape/chase rank distance and the ground truth, but we also expect the differences in rank distances to be large when the models are trained with low and high \emph{escape/chase fraction}. 

In order to quantify these differences, we computed the linear regression of \\
$log_{10}(rank ~ distance)$ vs the \textit{escape/chase fraction}, and we used the log-linear slope as a measurement of performance.  Thus, the more negative the slope is, the better the performance. It should be noted that in this analysis the series in which the word \textit{run} appears less that 5 times were excluded.

\subsection{Corpora}
In both test, we use as training corpora the TASA corpus \cite{TASA} and a random subsample of ukWaC corpus \cite{wacky}. TASA corpus is a commonly used linguistic corpus consisting of 37k educational texts 
with a corpus size of 5M words in its cleaned form. UkWaC consists of web pages material from .uk domain. The random subsample has 140k documents with a corpus size of 57M words in its cleaned form.

For the case study we use the Dreambank reports corpus \cite{dreambank,Domhoff2008}. The DreamBank corpus consists of 19k dreams reports from 59 subjects, containing about 1.3M words in its cleaned form. 
 
To clean the corpora, we performed a word tokenization, discarding punctuation marks and symbols. 
Then, we transformed each word to lowercase and eliminated stopwords, using the stoplist in NLTK Python package \cite{Bird2009}. Also, all numbers were replaced with the string ``NUM''.

\section{Results}
\subsection{Corpus size analysis in the clustering test}
As a first step for all analyses, we carried out the Skip-gram parameter optimization for both tests (Table \ref{tab:w2v_params}). The best scores where selected to perform the corpus size analysis. In the semantic categorization test, in the case of TASA corpus, neg = 15 was chosen given that it showed slightly better performance. 

\begin{table}[ht]
\begin{center}
  \begin{tabular}{|c|c|c|c|c|c|c}
\cline{3-6}
 \multicolumn{1}{} & & & win \textbackslash neg & 5 & 10 & 15 \\ \cline{1-6}
 \multirow{6}{*}{\rotatebox{90}{Silhouette score} }&\multirow{3}{*}{\rotatebox{90}{TASA}}  & 5 & 0.107 & 0.107 & 0.109     \\ \cline{3-6}
     &                     &  10 & 0.110 & 0.117 & 0.119      \\ \cline{3-6}
    &     		        &  15 & 0.115 & \textbf{0.121} & \textbf{0.121}\\ \cline{2-6}
 & \multirow{3}{*}{\rotatebox{90}{ukWaC}}  & 5 & 0.150 & 0.151 & \textbf{0.155}  \\ \cline{3-6}
  &                        &  10 & 0.146 & 0.149 & 0.151 \\ \cline{3-6}
   &      		        &  15 & 0.141 & 0.145 & 0.145      \\ \cline{1-6}
   \multirow{6}{*}{\rotatebox{90}{Correlation}} & \multirow{3}{*}{\rotatebox{90}{TASA}}  & 5 & 0.603 & 0.592 & 0.589     \\ \cline{3-6}
     &                     &  10 & 0.615 & 0.610 & 0.602      \\ \cline{3-6}
    &     		        &  15 & 0.623 & 0.618 & \textbf{0.626} \\ \cline{2-6}
 & \multirow{3}{*}{\rotatebox{90}{ukWaC}}  & 5 & 0.643 & 0.633 & 0.638   \\ \cline{3-6}
  &                        &  10 & 0.644 & 0.643 & 0.637  \\ \cline{3-6}
   &      		        &  15 & \textbf{0.647} & 0.640 & 0.642       \\ \cline{1-6}
  \end{tabular}
\end{center}
\caption{\label{tab:w2v_params} Skip-gram's parameter selection. Silhouette scores for the categorization test and correlations for the WordSim353 test. In all cases the embedding dimensions were set to 100.}
\end{table}

\begin{figure}[H]
\begin{center}
  \includegraphics[width=0.98\textwidth]{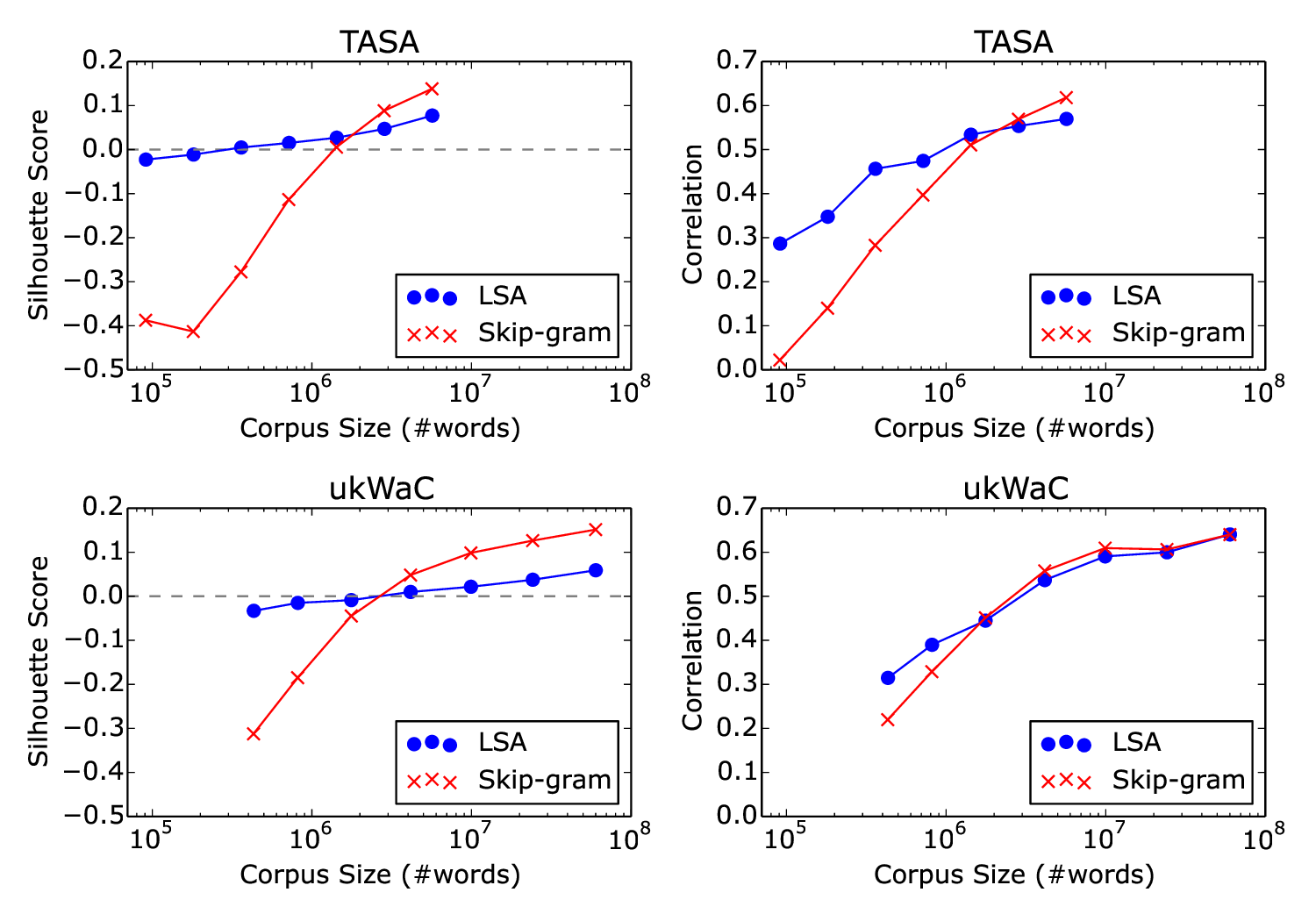}
\caption{ Semantic categorization test performance (left graphs) and WordSim353 test performance (right graphs) in function of the corpus size for LSA and Skip-gram model. The size of the different corpus are considered as the number of tokens that they contain.}
\label{subcorpus_analysis}
\end{center}
\end{figure}

To compare LSA and Skip-gram embeddings quality in small size corpora, we tested both methods in random nested subsamples of TASA and ukWaC corpus (see Figure \ref{subcorpus_analysis}).
Given that the appropriate embeddings dimensions depends on the corpus size \cite{fernandes2011automatic}, for each sub-corpus, we ran the models with a wide range of dimension values (7,15,25,50,100,200,400), using in each case the dimension that produces the best performance.

Figure \ref{subcorpus_analysis} shows that Skip-gram word-knowledge acquisition rate tends to be larger than LSA's. While Skip-gram tends to produce better embeddings than LSA when they are trained with large corpora, under training with small corpora Skip-gram performance is considerable lower than LSA's.
We believe that this behavior is grounded in the fact that Skip-grams is a prediction-based model, so it requires substantial training data in order to fit its high number or parameters.

\subsection{Case study: semantic association in dreams report}\label{results_dreams}
In order to check the expected differences between the associations of the word \textit{run} in dreams and waking life,
we built LSA and Skip-gram embeddings trained each corpora, and we extracted the 25 words most similar to \textit{run} in each case. Infrequent words which appear less than 15 times were excluded. We found that word embeddings are capable of identifying differences in usage patterns of word between dreams and waking life. In TASA and ukWaC corpora, \textit{run} is linked with words associated with a big variety of contexts, such as sports, means of transport and programming, while in dreams, \textit{run} is directly related with words associated with chase/escape situations. For example with LSA trained in dreams we obtained words such as: \textit{chase, scream, chasing, escape, chases, grab, screaming, nazi, hide, chased, yells, safety, devil, evil, attacking, killing, slam and yell}. In the same line, with Skip-gram model we found words such as: \textit{escape, catch, chase, chasing, follow, dangerous, guards, robbers, hide, hiding, escaped, safely, safe, protect and tornado}.

Then, we tested the ability of both models to extract semantic tendencies in single dreams series following the method described in subsection \ref{method_dreams}. A parameter selection was made, obtaining the best performance for LSA in 200 dimensions and for Skip-gram in win=15 and neg=10 (Table \ref{tab:SG_params_dreams} and Table \ref{tab:LSA_params_dreams}).

\begin{table}[ht]
\begin{center}
  \begin{tabular}{|c|c|c|c|}
  \hline	
  win \textbackslash neg & 5 & 10 & 15 \\ \hline	
   5 & -0.63 & -0.96 & -0.99     \\ \hline	
  10 & -0.95 & -1.06 & -1.01      \\ \hline	
  15 & -0.95 & \textbf{-1.12} & -1.02   \\ \hline
  \end{tabular}
\end{center}
\caption{\label{tab:SG_params_dreams} Skip-gram's parameter selection in the dreams reports analysis. The scores are the slopes in the log-linear regression of the escape/chase rank distance vs escape/chase fraction. The shown values are the mean score among 10 repetitions.The embedding dimensions were set to 50.}
\end{table}

\begin{figure}[H]
\begin{center}
  \includegraphics[width=0.6\textwidth]{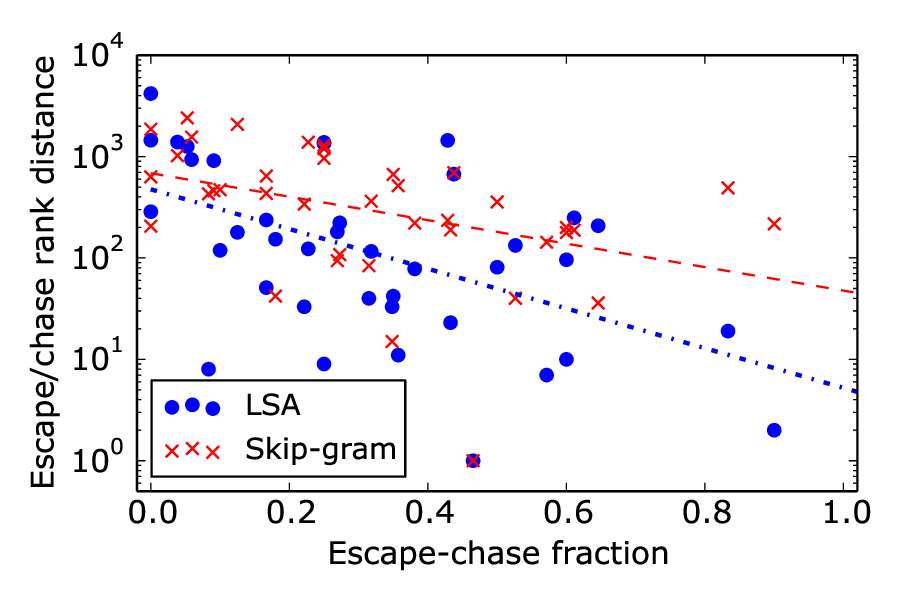}
\caption{Escape/chase rank distance vs the escape/chase fraction for each individual series. A log-linear regression was performed for one sample of LSA and Skip-gram models (blue dash-dotted line and red dashed line respectively). LSA measurements present a log-linear slope of -2.10, while the Skip-gram model has a slope of -1.11. Also, LSA and Skip-gram model show correlation of -0.57 and -0.42, with \emph{p-values} of 0.0001 and 0.007 respectively.}
\label{minrank_dreams}
\end{center}
\end{figure}

\begin{table}[ht]
\begin{center}
{
  \begin{tabular}{|c|c|c|c|c|c|c|}
  \hline	
  dim & 30 & 50 & 100 &200 & 300 & 400  \\ \hline	
  Slope & -1.65 & -1.67 & -1.96 & \textbf{-1.99} & -1.73 & -1.77     \\ \hline
  \end{tabular}
 }
\end{center}
\caption{\label{tab:LSA_params_dreams} LSA's parameter selection in the dreams reports analysis. The scores are the slopes in the log-linear regression of the escape/chase rank distance vs escape/chase fraction. The shown values are the mean score among 10 repetitions}
\end{table}
The sensitivity of the method to detect both situation of the term \emph{run} rely on the slope steepness.
We expect negative correlations with steep slopes between the escape/chase rank distance and the escape/chase fraction (see Methods section for details). 
In figure \ref{minrank_dreams} we plot the calculated distance vs the ground truth for each individual series in the selected parameters.

While both models present a downward trend, the LSA outperforms Skip-gram with a negative log-linear slope of -1.99 and -1.12, respectively.
We ran this test for 10 iterations and slopes showed significant difference between methods (Kolmogorov-Smirnoff test, $p<3\times10^{-4}$).

\begin{figure}[H]
\begin{center}
\includegraphics[width=0.98\textwidth]{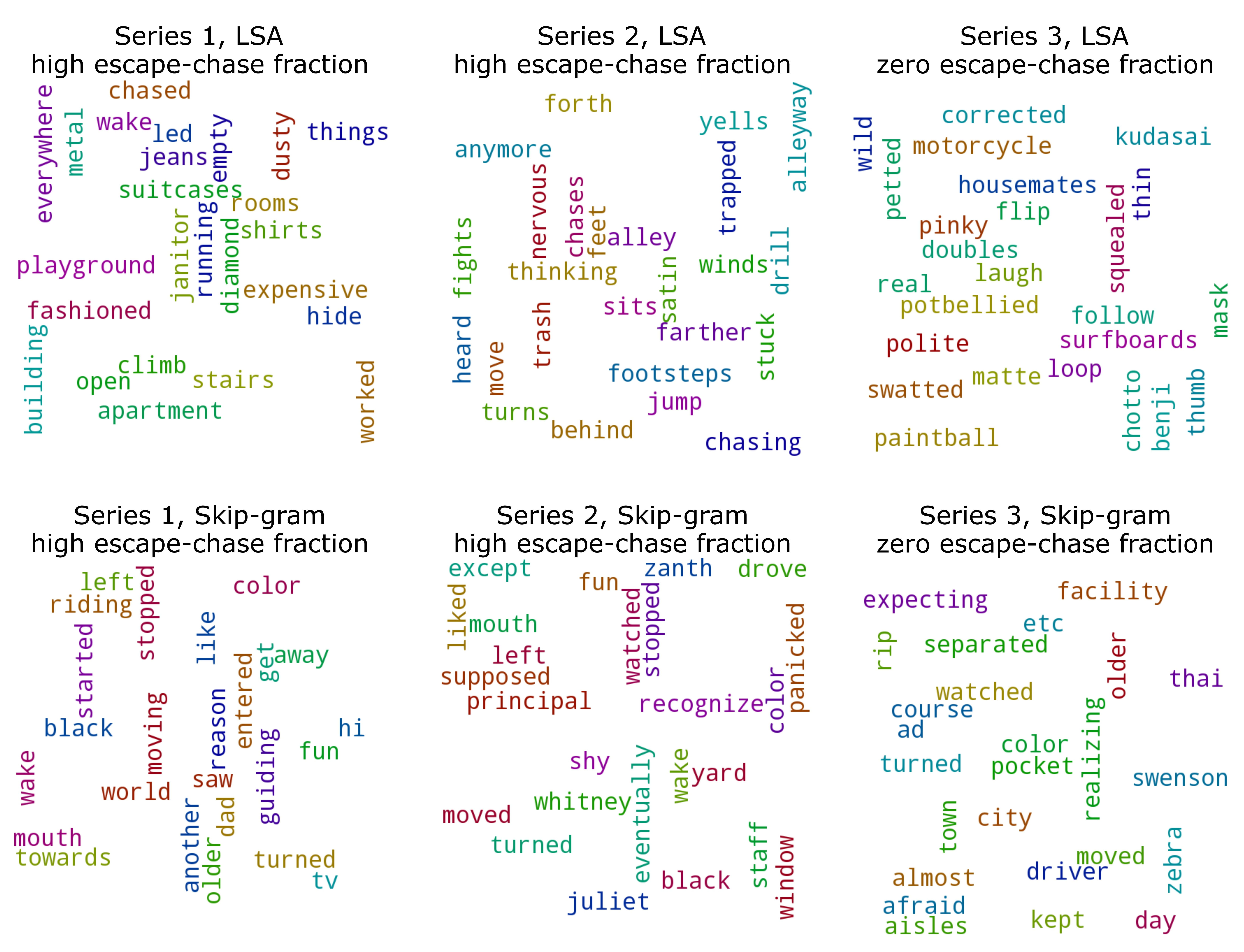}
\caption{ Semantic neighborhood of the word \textit{run} in 3 different dreams series for LSA and Skip-gram's words embeddings. Series 1 and 2 are the two series with the highest escape/chase fraction, while series 3 is a control series, which has no escape/chase situations in dreams that contain the word \textit{run}. Series 1 is the \textit{``seventh grade girls''} series, in which only 5 of its 69 dreams reports contain the word \textit{run} and on average 90\% of these dreams refer to chase/escape situations (escape/chase fraction of 0.9).  Series 2 is the \textit{``Bay Area girls: Grades 7-9''} series, in which 6 of its 154 dreams reports contain the word \textit{run}, 83\% of which refer to a chase/escape situation (escape/chase fraction of 0.833). Series 3 is the \textit{``Madeline3: Off-Campus''} series, in which 13 of its 348 dreams reports contain the word \textit{run} and none of them refers to a chase/escape situation (escape/chase fraction of 0).}
\label{wordclowd_fig2}
\end{center}
\end{figure}

In order to illustrate to what extent we can use these methods to explore the usage pattern of a target word in individual dream series, we show in figure \ref{wordclowd_fig2} the 25 closest words of \textit{run} in 3 different dreams series, using the same parameter set as in figure \ref{minrank_dreams}. Series 1 and 2, are the two series with the highest escape/chase fraction, while series 3 has no escape/chase situations in dreams that contain the word \textit{run}. In the first two series, we observe that \textit{run} neighborhood in LSA embedding contains words highly related with escape/chase situations, such as \textit{chased} and \textit{hide} in series 1 and \textit{chasing}, \textit{chases} or \textit{trapped} in series 2.
Conversely, Skip-gram embeddings do not succeed in identifying escape/chase contexts in these series. 
As a control case, it can be seen that series 3 do not show escape/chase related words. 

\subsection{Conclusion}

In the present paper, we compare the capabilities of Skip-gram and LSA to learn accurate word embeddings in small text corpora. In order to do that, we first tested the models capability to represent semantic categories (such as drinks, countries, tools or clothes) in nested subsamples of a medium size corpus. We found that Word2vec embeddings outperform LSA's when the models are trained with medium size datasets ($\sim$ 10 millions of words). However, when the corpus size is reduced, Word2vec performance has a severe decrease, thus LSA becoming the more suitable tool. This finding gives a new insight into the prediction-based vs counter-based models discussion \cite{Baroni2014,Levy2014,Levy2015}. We believe that Word2vec performance decrease in small corpora is grounded in the fact that prediction-based models need a lot of training data in order to fit their high number of parameters. 

As a case study, we have studied LSA and Skip-gram capabilities to extract relevant semantic words associations in dreams reports. We found that LSA can accurately capture semantic words relations even in cases of series with low number of dreams and low frequency of target words. This is a step foward to the application of word embeddings to the analysis of dreams content. This research field addresses questions such as ``what do we dream about?'' and ``how do gender, cultural background and waking life experiences shape the dreams content?'' \cite{bell2011personality,domhoff2000methods,Domhoff2002,Domhoff2008}.
We propose that LSA can be used to explore words associations in dreams reports, which could bring new insight into this old research area of psychology.

\section*{Acknowledgments}
We want to thank the teams behind the TASA \cite{TASA}, WaCky \cite{wacky} and Dreambank \cite{Domhoff2008} projects for providing us the corpora and Eduardo Schmidt for helpful discussions.

\section*{Conflict of Interest Statement}
The authors declare that there is no conflict of interest regarding the publication of this paper.

\bibliographystyle{unsrt}
\bibliography{altszyler_et_al}

\end{document}